\documentclass[conference]{IEEEtran}
\ifCLASSINFOpdf
\else
\fi
\usepackage{graphicx}
\usepackage{amsmath, amssymb}
\usepackage{bm} 
\usepackage{gensymb} 

\newcommand{\mat}[1]{\boldsymbol{#1}}
\newcommand{\Gparam}{\theta_G}
\newcommand{\Dparam}{\theta_D}
\newcommand{\oriImg}{\mat{X}}
\newcommand{\corImg}{\mat{Z}}
\newcommand{\recImg}{\hat{\mat{Z}}}
\newcommand{\Gtrainedparam}{\hat{\theta}_G}
\newcommand{\mask}{\mat{M}}
\newcommand{\fourier}{\mat{F}}
 
\newcommand{\contentLoss}[2]{\ell_{\text{content}}(G_{\Gparam}(#1), #2)} 
\newcommand{\advLoss}[1]{\ell_{\text{adversarial}}(G_{\Gparam}(#1))} 

\DeclareMathOperator*{\minimize}{minimize}

\begin{document}
%
\title{Generative Adversarial Networks for Recovering Missing Spectral Information}

\author{\IEEEauthorblockN{Dung N. Tran, Trac D. Tran}
\IEEEauthorblockA{Department of Electrical and\\Computer Engineering\\
Johns Hopkins University\\
Baltimore, MD 21218\\
Email: dung.n.tran@jhu.edu, trac@jhu.edu}
\and
\IEEEauthorblockN{Lam Nguyen}
\IEEEauthorblockA{U.S. Army Research Laboratory\\
2800 Powder Mill Rd\\
Adelphi, MD 20783\\
Email: Lam.H.Nguyen2.civ@mail.mil}}


%


\maketitle

\begin{abstract}
Ultra-wideband (UWB) radar systems nowadays typical operate in the low-frequency spectrum to achieve penetration capability. However, this spectrum is also shared by many others communication systems, which causes missing information in the frequency bands. To recover this missing spectral information, we propose a generative adversarial network, called SARGAN, that learns the relationship between original and missing band signals by observing these training pairs in a clever way. Initial results shows that this approach is promising in tackling this challenging missing band problem.
\end{abstract}


%
\IEEEpeerreviewmaketitle

\section{Introduction}


 
Over the past few decades, ultra-wideband (UWB) radar systems have been widely employed in various practical applications due to their penetration capability. For example, the U.S. Army has been developing UWB radar systems for detection of difficult targets in various applications such as foliage penetration \cite{Nguyen1997}, ground penetration \cite{Nguyen1998}, and sensing-through- the-wall \cite{Nguyen2008}. To achieve penetration capability, these systems must operate in the low-frequency spectrum that spans from under $100$ MHz to several GHz. In addition to the low-frequency requirement for penetration, they must employ wide-bandwidth signals to achieve the desired resolution. However, the signal occupies a wide spectrum that is also shared by radio, TV, cellular phones, and other systems. The frequency allocation and use problem thus becomes a major challenge and only worsens over time as additional radar and communication systems that need the penetration feature must operate in this low-frequency spectral region.

There are two key challenges for any UWB system: 1) the system must operate in the presence of other systems and 2) the system must avoid transmitting energy in certain frequency bands that are specified by frequency management agencies. As a result, the receive data have a spectral content that includes multiple bands that are either corrupted (due to the presence of interference sources) or nonexistent (because of no transmission in the prohibited frequency bands). In this paper, we tackle the latter problem in which a large portion of the spectrum is notched due to the frequency allocation issue.

Conventional techniques usually detect the corrupted frequency bands by searching for spikes in the spectral domain. The fast Fourier transform (FFT) bins that correspond to the contaminated frequency bands are zeroed out. This technique results in severe sidelobes in the time or spatial domains of the output data and imagery due to the sharp transitions (frequency samples with no information) in the frequency domain. To overcome these limitations, Do et. al. \cite{Nguyen2012} proposed a technique to recover missing spectral information using sparse representation. It is based on the assumption that the full spectrum data and corrupted versions are similarly sparsely represented by a full spectrum dictionary and a missing band dictionary, respectively. Its limitation is a lack in the ability to distinguish near-by targets at fine resolution. Furthermore, the missing frequency bands are required a priori.

Recently, a class of generative model in neural network literature, namely, Generative Adversarial Network (GAN) \cite{Goodfellow2014}, has produced remarkable results in various applications in computer vision, speech processing, and other fields. A standard GAN takes a random noise vector as an input and generates samples that resemble real data. There are also
many works that feed GAN with conditions, such that the
generated image samples are not only realistic but also
match the constraints imposed by the conditions. Some
works conditioned GAN on discrete class labels \cite{Chen2016, Mirza2014},
while many other works synthesized images by conditioning
GAN on images for the tasks such as domain transfer
\cite{Yoo2016, Isola2017}, image super-resolution \cite{Johnson2016, Ledig2017},
image synthesis from surface normal maps \cite{Wang2016}, and style
transfer \cite{Li2016}.

In this paper, we propose a GAN framework to recover missing spectral 
information in multiple frequency bands of UWB synthetic aperture radar (SAR) data that are 
either corrupted or nonexistent. Specifically, we propose a generator loss function that encourages the network to
seek solutions on the SAR image manifold that are consistent with data in the frequency domain. Our proposed
method can be seen as a variant of a conditional GAN framework,
but conditioned on the spectral domain.

The network is trained by observing various spectral missing patterns. The advantage of this technique
is twofold. First, all computational complexity is at the training phase. 
The testing phase only consists of some simple matrix multiplication. 
Second, to recover a SAR image from its frequency corrupted version, the trained network requires zero 
information of the missing band locations. This is an advantage of our proposed method over traditional spectral recovery techniques in which missing frequencies are required a priori. To our knowledge, this is the first GAN-based framework for recovering missing spectral information in UWB radar systems.

\section{Method}
We aim to reconstruct a SAR image $\oriImg$ from its missing band version 
$\corImg$. In our framework, we adopt a GAN structure. We train the network by minimizing a standard discriminator loss and a 
generator loss specifically designed for this missing spectral problem. The training data include 
a set of image pairs, each consisting of an uncorrupted image and its frequency-corrupted counterpart.
Each corrupted image is obtained by notching out certain frequency bands of
the original image. Original images are not available in the testing phase.

Our goal is to train a generator $G_{\Gparam}$, parameterized by $\Gparam$ that reconstructs a SAR image from its frequency-corrupted version. Given a set of training data $\{(\oriImg_j, \corImg_j)\}_{j=1}^n$, we train the generator by solving
\begin{equation}
	\displaystyle{\minimize_{\Gparam} \sum_{j=1}^n \mathcal{L}(G_{\Gparam}(\corImg_j), \oriImg_j)}.
\end{equation}
Then a SAR image can be recovered from its missing band counterpart $\corImg$ as
\begin{equation}
	\recImg = G_{\Gtrainedparam}(\corImg).
\end{equation}
We describe our generator loss in detail in Section \ref{subsec:gen_loss}. It conditions on the frequency domain of the generated sample and forces the generator to favor solutions on the SAR image manifold.

\subsection{Generative Adversarial Networks}
GANs are neural networks
for training generative models in an adversarial manner. 
A GAN consists of two networks, a generator $G$ and a discriminator $D$. 
The generative network $G$ learns a mapping from a low-dimensional representation space to a high-dimensional space.  The purpose of $G$ is to generate samples that resemble the training data.
The discriminator $D$ maps an input to a likelihood. Its role is to distinguish between the sample generated 
by $G$ and the sample from the data distribution.

Directly applying standard GANs to the missing spectral recovery problem fails to reconstruct original images, as they produce samples that 
are inconsistent with the input data in the frequency domain. We therefore formulate our generator loss to favor 
solutions that contain available frequencies in the corrupted images. This guarantees consistency between the generated sample and the original image. Moreover, the input in our generator is a corrupted image instead of a low-dimensional encoding as in traditional GANs. This allows our network to learn a mapping from a corrupted input to a desired solution.

\begin{figure}[h]
  \centering
  \includegraphics[width=0.48\textwidth]{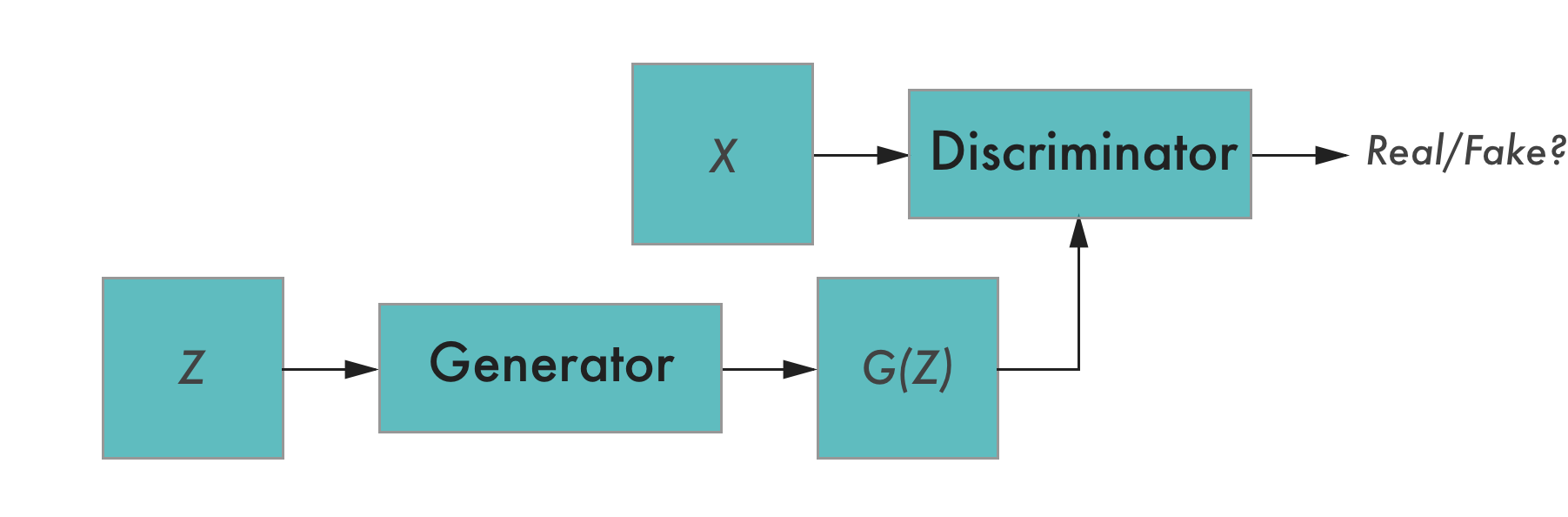}
  \caption{SARGAN architecture. The generator produces an estimate of a full spectrum image from its corrupted version to full the discriminator. The discriminator tries to distinguish this estimate with the original image. In a successfully trained SARGAN, the generator produces estimates that are close to the full spectrum image, thus successfully fool the generator.}
  \label{fig:sargan}
\end{figure}

\subsection{Generator Loss}\label{subsec:gen_loss}
We encourage the generator to seek for solutions on the SAR image manifold that are consistent with the input. To do so, we formulate the generator loss as a weighted sum of a content loss component and an adversarial loss component.

A SAR image and its missing band counterpart is related by:
	\begin{equation}
		\fourier \corImg = \mask \circ \fourier \oriImg.
	\end{equation}
Here, $\fourier$ is the Fourier matrix, and $\mask$ is a binary masking matrix defined as:
\begin{equation}
	\mask_{i,j} = \begin{cases} 
   1 & \text{if } (i,j) \text{ is in an available band}, \\
   0       & \text{if } (i,j) \text{ is in an notched band}.
  \end{cases}
\end{equation}
In other words, the masking matrix notches out missing frequency bands and preserves the available frequencies in the original image. As missing local information in the frequency domain results in a global deviation in the time domain, imposing data consistency in the time domain fails to recover notched spectral information. We therefore define a content loss that requires generated samples to preserve available frequencies in the input images:
\begin{equation}
	\contentLoss{\corImg}{\oriImg} = \big\|\mask \circ \fourier G_{\Gparam}(\corImg) - \mask \circ \fourier \oriImg\big\|_1,
\end{equation}
where the $L_1$ loss is defined as $\|\boldsymbol{A}\|_1 = \sum_{i,j} |\boldsymbol{A}_{i,j}|$, for a given matrix $\boldsymbol{A}$. Note that the $L_1$ loss can be replaced by other losses such as $L_2$. In our experiments, we find that the $L_1$ loss results in a faster convergent rate and more robust reconstruction than the $L_2$ loss.

To further improve the reconstruction quality, we impose an adversarial loss to the generator. This encourages the generator to fool the discriminator by seeking solutions on the SAR image manifold:
\begin{equation}
	\advLoss{\corImg} = - \log D_{\Dparam} \left( G_{\Gparam} (\corImg) \right)
\end{equation}
The generator loss is defined as a weighted sum of these two losses:
\begin{equation}
	\mathcal{L}(G_{\Gparam}(\corImg), \oriImg)  = \contentLoss{\corImg}{\oriImg} + \lambda \advLoss{\corImg},
\end{equation}
where $\lambda > 0$ is a positive constant controlling the tradeoff between the two terms.
\subsection{Discriminator Loss}\label{subsec:dis_loss}
We adopt a standard discriminator network $D_{\Dparam}$ which we train to solve the following optimization problem:
	\begin{align*}
		\max_{\Dparam} & \;\mathbf{E}_{\oriImg \sim p_{\text{data}}(\oriImg)}\left[\log D_{\Dparam}(\oriImg)\right] + \\
		&\;\mathbf{E}_{\corImg \sim p_{G}(\corImg)}\left[1 - D_{\Dparam}\left(G_{\Gparam}(\corImg)\right)\right].
	\end{align*}
This allows one to train a generator to produce realistic SAR images from corrupted inputs to fool a discriminator, which is trained to differentiate reconstructed SAR images from original ones. Our generator is thus encouraged to favor solutions on the SAR image manifold.

\section{Results}
	In this section, we demonstrate SARGAN for the spectral recovery problem using SAR data from the U.S. Army Research Laboratory (ARL) UWB SAR system. 
	
\begin{figure}[!htb]
    \centering
    \begin{minipage}{.15\textwidth}
        \centering
        \includegraphics[width=\textwidth]{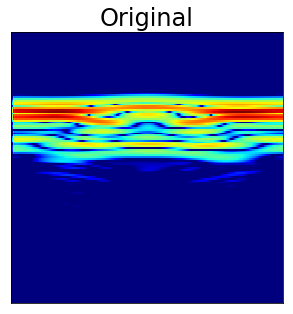}
    \end{minipage}%
    \begin{minipage}{.15\textwidth}
        \centering
        \includegraphics[width=\textwidth]{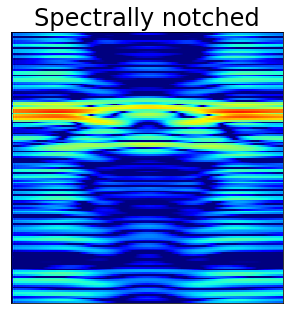}
    \end{minipage}
    \begin{minipage}{.15\textwidth}
        \centering
        \includegraphics[width=\textwidth]{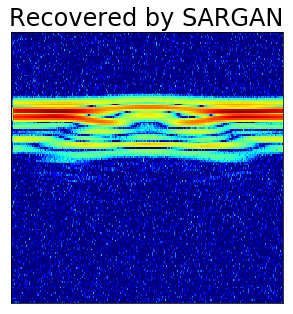}        
    \end{minipage}
    \caption{Raw data in time domain of target versus aspect angle. }
            \label{fig:raw_data}
\end{figure}

	This SAR database consists of targets (metal and plastic mines, 155-mm unexploded ordinance [UXO], etc.) and clutter objects (a soda can, rocks, etc.) buried under rough ground surfaces. The electromagnetic (EM) radar data are simulated based on the full-wave computational EM method known as finite-difference, time-domain (FDTD) software [26], which was developed by ARL. The software was validated for a wide variety of radar signature calculation scenarios [27], [28]. Our volumetric rough ground surface grid with the embedded buried targets was generated by using the surface root-mean-square (rms) height and the correlation length parameters. The targets are flush buried at a 2-3 cm depth. Fig.~\ref{fig:raw_data} (left) shows original SAR raw data (using VV polarization) of some targets that are buried under a perfectly smooth ground surface. Each target is imaged at a random viewing aspect angle and an integration angle of $60\degree$.
	
	In our experiment, the SAR radar is configured in side-looking mode. It travels in the horizontal direction, transmits impulses to the imaging area, and receives backscattered radar signals from the targets. In this scene, there might be many point targets that have different amplitudes and are located randomly throughout the scene. For demonstrating purposes, we use the raw data in a case where there is a random point target on the scene. The left image in Fig.~\ref{fig:raw_data} shows the full spectrum raw data for this simulation scenario. The data bandwidth is from $380$ MHz to $2.08$ GHz, which contains $90\%$ of the signal energy. It serves as the baseline image for performance comparison purposes.

\begin{figure}[h]
  \centering
  \includegraphics[width=0.48\textwidth]{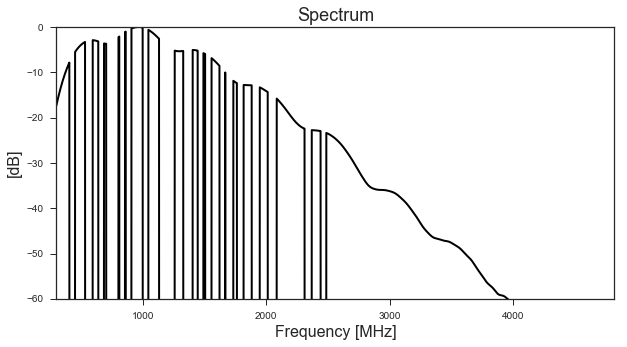}
  \caption{Spectrum of the raw data with 90\% missing in the bandwidth.}
  \label{fig:spectrum}
\end{figure}

	Next, we consider the spectral notches case due to the frequency allocation restriction. In our experiments, we randomly zero out frequency sub-bands of the spectrum, each equivalent to $10$ times the frequency resolution which is equal to $9.15$ MHz. These random frequency bands can be overlapped and sum up to $90\%$ of the data spectrum. Fig.~\ref{fig:spectrum} demonstrates the aforementioned randomly notching procedure in the frequency domain of the data.
	
\begin{figure}[h]
  \centering
  \includegraphics[width=0.48\textwidth]{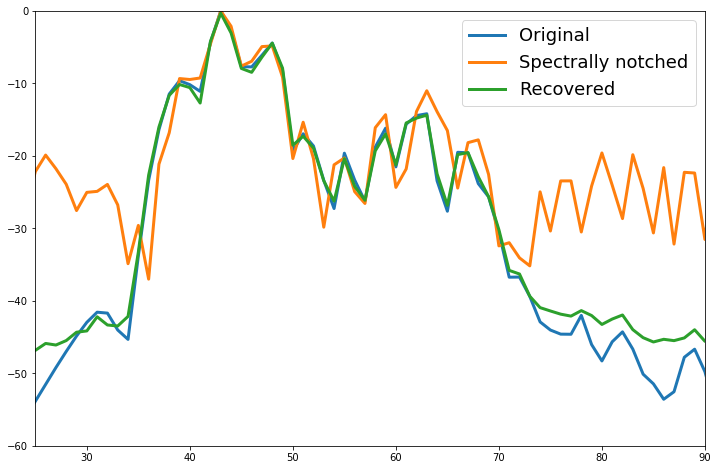}
  \caption{Normalized down-range profiles in dB scale of the raw data. The spectrally notched data show severe sidelobes, whereas the data reconstructed using SARGAN follow the ground-truth very well. The test reconstructed result was obtained after $100$ training epochs.}
  \label{fig:downrange_profiles}
\end{figure}

	The middle image in Fig. ~\ref{fig:raw_data} shows the raw data with $90\%$ of the spectrum being notched. Fig.~\ref{fig:downrange_profiles} presents the downrange profiles of the data. The large amount of missing frequencies results in severe sidelobes in the data. Recovering the original data is therefore challenging in this situation.
	
	We use SARGAN to recover missing spectrum information under this setup. We use a four-layer fully connected neural network for the generator, and a three-layer fully connected neural network for the discriminator. The first and last layers of the generator have the same dimensions as the input data. The two hidden layers are of length $128$. The dimension of the first layer of the discriminator is equal to that of the input data. Its hidden layer has $128$ nodes and its output has one node, which guesses whether the input is a true image or one produced by the generator. In our experiments, we use a stable alternative to GANs, called Wasserstein GAN (WGAN). 
	
	The training data are obtained as follows. From a full spectrum raw data, we produce several randomly spectrally notched version of that data. Each such notched data matrix together with the full spectrum data constitute a training pair. We then train our network using these training pairs. In the testing phase, the full spectrum raw data are unavailable. Our goal is to recover it from a corrupted version that is not included in the training data. The locations of the notched band are unknown to the network. This is significantly different than other traditional spectral recovery techniques in which missing frequencies are required a priori. 
	
\begin{figure}[h]
  \centering
  \includegraphics[width=0.48\textwidth]{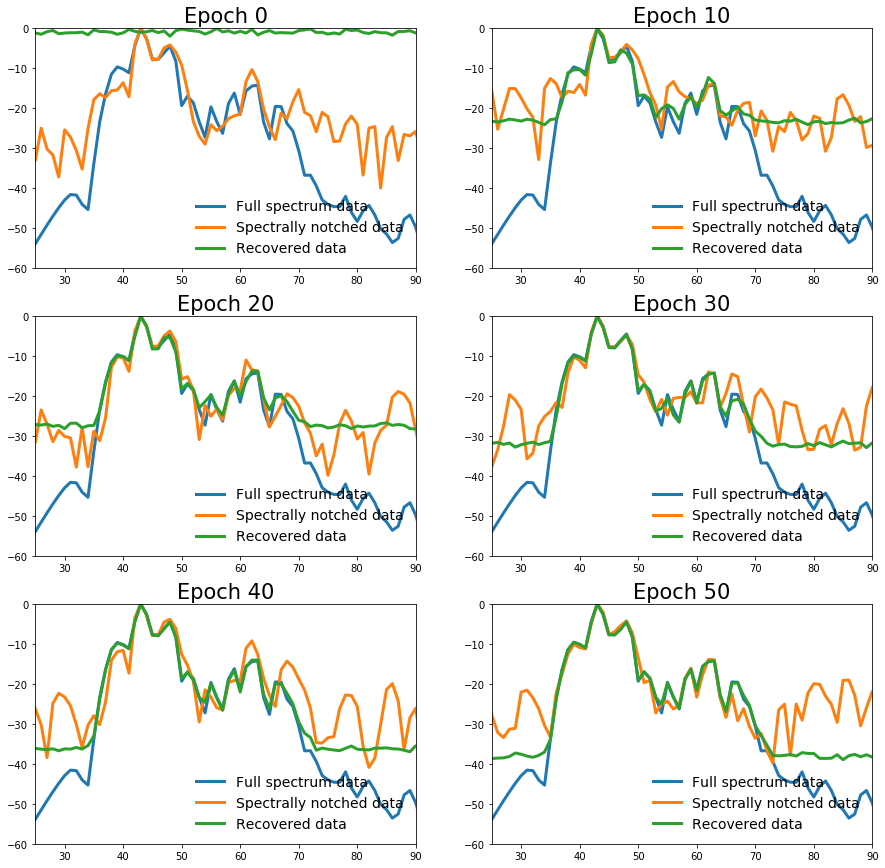}
  \caption{Normalized down-range profiles in dB of the reconstructed data during the first $60$ epochs. Each figure shows the testing result after each $10$ epochs.}
  \label{fig:recovery}
\end{figure}

	Fig.~\ref{fig:recovery} shows the normalized down-range profiles in dB of the recovered data, produced by the generator of SARGAN, after each $10$ epochs. It can be seen that after $40$ epochs, the generated samples already well approximate the original full spectrum data. The normalized downrange profile of the test reconstructed data using SARGAN after $100$ training epoches are shown in Fig.~\ref{fig:downrange_profiles}. The recovered data closely follow the original data whereas the corrupted version show severe sidelobes.

\begin{table}[!t]\label{table:snr}
\renewcommand{\arraystretch}{1.3}
\caption{Recovery Performance of SARGAN on Spectrally Notched Data  (SNR)}
\label{table_example}
\centering
\begin{tabular}{|c|c|c|}
\hline
Corrupted Data SNR (dB) & Recovery SNR (dB) & Recovery Gain (dB)\\
\hline
8.15 & \textbf{23.99} & \textbf{15.84}\\
\hline
\end{tabular}
\end{table}

\begin{figure}[h]
  \centering
  \includegraphics[width=0.48\textwidth]{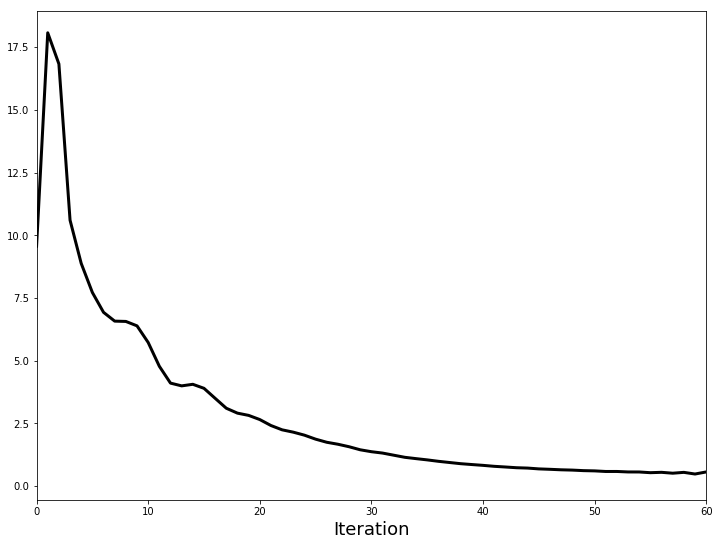}
  \caption{Generator loss values during the training phase. The network converges after roughly $60$ iterations.}
  \label{fig:generator_loss}
\end{figure}

Fig.~\ref{fig:generator_loss} visualizes the generator loss during the training phase. It can be seen that the network converges after around $50$ epochs, which matches the above down-range profile visualization.

To further qualitatively evaluate SARGAN, we compute the Signal-to-Noise Ratio (SNR) in dB scale between the original data $\oriImg$ and the data recovered using SARGAN $\recImg$ using the formula:
\begin{equation}
	\text{SNR}(\oriImg, \recImg) = 20\log_{10}\frac{\text{RMS}(\oriImg)}{\text{RMS}(\recImg - \oriImg)},
\end{equation}
where
\begin{equation}
	\text{RMS}(\oriImg) = \frac{1}{\sqrt{\text{\# of elements in }\oriImg}}\| \oriImg \|_2.
\end{equation}
We also compare that to the SNR between the original data and the corrupted data. Table~\ref{table:snr} shows the SNR in these two cases, and the performance gain in dB scale obtained by SARGAN. It can be seen that our proposed method reduced the sidelobe level by more than $15$ dB. This matches the normalized down-range profiles shown in Fig.~\ref{fig:downrange_profiles}. Remarkable, SARGAN obtains this performance gain without any information on the missing band locations. Popular methods such as FFT and sparse recovery fail in this case.

\section{Conclusion}
We proposed a Generative Adversarial Network framework, called SARGAN, to tackle the missing spectral information recovery problem. A well-trained SARGAN is expected to produce a good estimate of the full spectrum SAR data from its spectrally notched counterpart without any spectral information.  In the training phase, the network is encouraged to learn the relationship between a set of full and corrupted spectrum data pairs. This relationship is captured in our proposed generator loss function, which forces SARGAN to favor solutions on the SAR data manifold which are consistent with the input data in the frequency domain. 

Using the real UWB SAR database from database, we show that the proposed framework can successfully recover the information from the missing frequency bands. Remarkably, it obtains more than $15$ dB gain without knowing the missing frequency locations. To our knowledge, it is the first method obtaining such performance gain in this situation.






%

\end{document}